\documentclass[preprint,12pt,authoryear]{elsarticle}

\usepackage{amssymb}
\usepackage{algorithmic}
\usepackage{algorithm}
\usepackage{array}
\usepackage[caption=true,font=normalsize,labelfont=sf,textfont=sf]{subfig}
\usepackage{textcomp}
\usepackage{stfloats}
\usepackage{url}
\usepackage{verbatim}
\usepackage{graphicx}
\usepackage{cite}
\usepackage{booktabs}
\usepackage{bbm}
\usepackage{amsmath}
\usepackage{amssymb}
\usepackage{algorithm}
\usepackage{algorithmic}
\usepackage{color}

\usepackage{lettrine}

\newtheorem{proposition}{{Proposition}}
\newtheorem{proof}{{Proof}}
\usepackage{comment}

\journal{Pattern Recognition}

\begin{document}

\begin{frontmatter}



\title{Robust Graph Structure Learning under Heterophily}

\author[1]{Xuanting Xie}
\ead{x624361380@outlook.com}
\address[1]{{School of Computer Science and Engineering},
            {University of Electronic Science and Technology of China}, 
            {Chengdu},
           {611731}, 
            {China}}

\author[1]{Zhao Kang\corref{cor1}}

\cortext[cor1]{Corresponding author}

\ead{zkang@uestc.edu.cn}

\author[1]{Wenyu Chen}
\ead{cwy@uestc.edu.cn}    
\begin{abstract}
Graph is a fundamental mathematical structure in characterizing relations between different objects and has been widely used on various learning tasks. Most methods implicitly assume a given graph to be accurate and complete. However, real data is inevitably noisy and sparse, which will lead to inferior results. Despite the remarkable success of recent graph representation learning methods, they inherently presume that the graph is homophilic, and largely overlook heterophily, where most connected nodes are from different classes. In this regard, we propose a novel robust graph structure learning method to achieve a high-quality graph from heterophilic data for downstream tasks. We first apply a high-pass filter to make each node more distinctive from its neighbors by encoding structure information into the node features. Then, we learn a robust graph with an adaptive norm characterizing different levels of noise. Afterwards, we propose a novel regularizer to further refine the graph structure. Clustering and semi-supervised classification experiments on heterophilic graphs verify the effectiveness of our method. 
\end{abstract}



\begin{keyword}


Robustness, topology structure, contrastive learning, graph filtering, clustering
\end{keyword}

\end{frontmatter}

\section{Introduction}
Graph structure is ubiquitous in the real world and has shown to be a powerful tool representing complicated relations among objects. For instance, nodes in a social network could be a single person or an organization, and connections show different types of social ties; nodes in protein structures are different chemical elements, and connections show different combinations. Graph Neural Networks (GNNs) have been extensively investigated as a powerful method for modeling graph data. Most of GNNs, such as GCN \citep{kipf2016semi}, GAT \citep{velivckovic2018graph}, JKNet \citep{xu2018representation}, and GPRGNN \citep{chien2021adaptive}, adopt a massage-passing architecture, which aggregates the node's neighborhood, to learn 
compact node representations \citep{kipf2016semi,wang2017community}.


GNNs have achieved remarkable success in many tasks, such as node clustering and semi-supervised classification. Clustering divides nodes into different groups in an unsupervised way \citep{zhu2022collaborative}, while semi-supervised classification predicts node labels by exploiting a small set of labels. Despite the strong capacity of GNNs to capture the structural and feature information of graph data, their performance highly depends on the quality of the data. Due to the recursive aggregation scheme of GNNs, small noise will spread to neighborhoods, deteriorating overall embedding quality.

Real-world graphs are always noisy and sparse \citep{FGC, CGC}. For instance, fraudulent accounts purposely connect to real ones and always inject wrong information to the representations of other nodes. Recent research points out that an unnoticeable, deliberate perturbation in graph structure considerably jeopardizes the performance of GNNs \citep{zhang2020gnnguard}. Besides, the predefined graph structure is often incomplete and carries partial information of the target system. For example, the existing molecular graph overlooks the non-bonded interactions, hindering our understanding of the atom interactions. 

To solve the above issues, several methods have been proposed to jointly optimize GNNs and graph structure to improve the performance of downstream tasks. Bayesian GCNN \citep{zhang2019bayesian} proposes an iterative learning model and considers the observed graph as the realization from random graphs. Geom-GCN \citep{pei2019geom} bridges the disparity between the observed graph and the latent continuous space by network geometry and redefines the graph for aggregation. SLAPS \citep{fatemi2021slaps} updates the graph by recovering the masked features under the guidance of GNNs and can handle nodes that do not have supervision information. 
Pro-GNN \citep{jin2020graph} refines the graph structure by imposing a sparse and low-rank constraint, making it robust to adversarial attacks. IDGL \citep{chen2020iterative} is an iterative method with robust embeddings, where Dirichlet energy is used to smooth the signals. GEN \citep{wang2021graph} iteratively generates graphs by embedding learned from GNNs and estimates graph structure to boost semi-supervised learning tasks robustly. However, these methods are targeting GNNs and are not for general situations.

In many scenarios, an explicit graph is even not readily available due to various reasons, which hinders us to reveal the underlying relations between data points. Some researchers found that graph improves the performance of learning methods over those that do not use graph information \citep{lingam2021glam}. To employ graph-based machine learning methods, we have to build a graph. To this end, some graph construction methods have been developed. The construction of a k-nearest neighbor (kNN) graph is widely adopted due to its simplicity. Local structure methods calculate
the probabilities that two instances are neighbors. The self-expression method learns the graph by modeling each sample as a linear combination of others. These approaches are data-dependent, and noise and outliers can easily damage the performance \citep{wong2019clustering}.


Furthermore, heterophily is largely ignored in most work, where nodes of different classes are connected. For example, opposite sex is more likely to have connections in a social dating network; different elements are prone to establish connections in protein structures. Most existing work is based on an underlying assumption of strong homophily. Forcing nodes from different classes to pass messages incurs wrong representations and heterophily can be harmless to GNNs in some strict conditions \citep{ma2022homophily}. Therefore, learning a graph topology from heterophilic data is more challenging. Despite its great difficulty, this problem has largely been overlooked.

To fill this gap, we propose a novel Robust Graph Structure Learning under Heterophily (RGSL) method to learn a high-quality graph from raw data. RGSL contains two main steps in a completely unsupervised manner: the first step is to smooth the node features with a high-pass filter; the second step is to learn a robust graph by applying an adaptive norm and designing a novel regularizer. In conclusion, the main contributions are as follows:

\begin{itemize}
    \item{We propose to use a high-pass filter to improve the node features of the heterophilic graph. Graph filtering produces a more discriminative representation by encoding the topology structure information into features.}
    \item{We propose a robust graph structure learning method, which adaptively suits different levels of noise. Positive samples are dynamically chosen without data augmentations to further refine the graph structure.}
    \item{Our method is tested for clustering and semi-supervised learning tasks using heterophilic graph datasets. Interestingly, experimental results demonstrate that our straightforward approach surpasses current deep neural network methodologies.}
\end{itemize}

The remaining sections of this paper are structured as follows: in Section \ref{r}, we introduce some related methods based on the techniques we use. Then, Section \ref{proposed} introduces the details of our approach. Section \ref{experimental1} presents a series of clustering experiments with real-world datasets to evaluate our method. Section \ref{experimental2} shows the experiments on semi-supervised learning tasks, including node classification and classification with noisy edges, which demonstrate the great potential of our method for other applications. Finally, Section \ref{conclude} draws the conclusion.

\section{Related Work} \label{r}
\subsection{Graph Learning}
Graph learning aims to achieve a high quality graph from feature data or raw graph. There are two main approaches in the literature. One is based on the idea of self-expression, where the combination coefficient characterizes the similarity between samples \citep{li2023explicit}. A number of regularizers, including the sparse norm, nuclear norm, Frobenius norm, and nonconvex norm \citep{zhang2023efficient,zhao2023self}, are applied to achieve the specific structure property of the graph. They are mainly used to reduce the noise. However, these regularizers often fail to fully express the rich topological priors \citep{pu2021learning}. 
Another line of research is based on the idea of adaptive neighbor. Contrary to the above approach, this method is supposed to capture the local structure. Numerous variants have been proposed \citep{wang2020gmc}. However, this kind of method highly depend on the assumption of data distribution. 

Recently, some methods have attempted to obtain a refined graph from the given graph. For example, FGC \citep{FGC} and GloGNN \citep{li2022finding} assume that the learned graph should be close to the structure information in the original graph by defining high-order relation. MCGC \citep{pan2021multi} uses a graph-level contrastive regularizer to improve the discriminability of the graph. Different from the popular contrastive mechanism, MCGC selects the nearest neighbors as positive samples. \citep{liang2019consistency} learns a clean graph from multiple noise graphs. They are still limited to some traditional scenarios.

Inspired by the success of GNNs, many models focus on the graph quality issue when applying GNNs. They either derive edge weights based on node representations or optimize the adjacency matrix and GNNs parameters jointly. These methods mainly pay attention to homophilic graphs and ignore the importance of heterophily in real world. Consequently, existing methods could produce inferior performance on heterophilic graph. Some methods try to learn a homophilic graph from a heterophilic one. For example, HOG-GCN \citep{wang2022powerful} learns a homophilic graph, in which the edge weights are the extent of nodes that belong to the same class over the whole heterophilic graph. In neighbor aggregation, intra-class nodes displaying lower homophily will exert a greater influence compared to the underlying inter-class nodes by incorporating class-aware information throughout the propagation. However, these methods rely on labels to select homophilic edges.

 Graph learning from heterophilic data in a completely unsupervised manner is urgently needed. To the best of our knowledge, CGC \citep{CGC} is the only shallow method that considers heterophily for graph clustering. CGC chooses the nearest neighbors in its contrastive learning mechanism and learns a new graph. However, it ignores the robustness of the model. Graph structures in real-world applications may undergo numerous perturbations, resulting in a significant decline in performance.

\subsection{Graph Filtering}
Recent research has found that GNNs only perform low-pass filtering on feature vectors and the graph helps to denoise the data. Inspired by this observation, some recent methods use graph filtering to preprocess the data without building on neural networks. AGC \citep{zhang2019attributed} applies a low-pass filter to process the node features and then performs spectral clustering on the smoothed data. MAGC \citep{lin2023multi} learns a graph from the smoothed multiview data by preserving high-order topology structures. Rather than using a fixed filter, MCGC \citep{pan2021multi} introduces a tunable parameter to the filter, so that it well suits different types of data. These works demonstrate that graph filtering is an effective way to improve data representation even without using deep neural networks. However, they ignore the fact that practical data could be heterophilic, where the underlying assumption of low-pass filtering is violated, i.e., low-frequency signals predominate over high-frequency ones. In heterophilic graph, linked nodes have different labels. Forcing nodes from different classes to pass messages blurs the distinction between their representations and results in sub-optimal performance. Therefore, we propose to use a high-pass filter in this paper.



\subsection{Contrastive Learning}
Contrastive learning is a popular self-supervised representation learning method, which is applied to make positive pairs close while negative samples apart, and has achieved great success in many tasks \citep{liu2022graph}. SimCLR \citep{chen2020simple} defines the pairs of positive and negative samples based on image augmentation. \citep{kong2019mutual} maximizes the mutual information between global sentence representations and n-grams in sentences. GraphCL \citep{you2020graph} designs different types of graph augmentation strategies for graph representation learning. GCA \citep{zhu2021graph} develops a data augmentation technique that is adaptive to graph structure and attributes. DCRN \citep{liu2022deep} uses distortions on the graph to obtain an augmented view. However, these techniques perform data augmentation using a fixed input graph, the quality of which will impact the augmentation quality. Moreover, different augmentation schemes may result in semantic drift and degrade the performance \citep{trivedi2022augmentations}. Therefore, some researchers have developed an augmentation-free self-supervised learning framework for graphs. For instance, \citep{lee2022augmentation} generates an augmented view by finding nodes that share the local structural and the global semantics information of the graph; \citep{dwibedi2021little} chooses nearest-neighbors as positive samples for ImageNet classification task. In this work, we design a novel regularizer to increase the discriminability of learned graphs using an augmentation-free method.

\section{Methodology} \label{proposed}
\subsection{Notation}
An undirected graph $\mathcal{G}=(\mathcal{V},E,X)$ is defined, where $\mathcal{V}$ represents a set comprising $N$ nodes, $X=\lbrace x_1,...,x_N\rbrace^{\top}$ is the feature matrix, and $ E $ is the set of edges. Given an adjacency matrix $A\in \mathbbm{R}^{N \times N}$, in which ${a}_{ij}=1$ if there is an edge between node $i$ and node $j$ and otherwise ${a}_{ij}=0$. $D=({A} + I)\mathbf{1}_{N}$ represents the degree matrix with a self-loop, where $I$ denotes the identity matrix. The normalized Laplacian graph is expressed as $ L = I - D^{-\frac{1}{2}}(A+I)D^{-\frac{1}{2}} $.

\subsection{Graph Filtering}
Considering the graph signal perspective, $X\in \mathbbm{R}^{N \times d}$ representing $N$ nodes can be understood as comprising $d$ $N$-dimensional graph signals. $L=U{\Lambda}U^\top$, where $\Lambda$ = diag[$\lambda_1$, $\lambda_2$, ..., $\lambda_N$] are the eigenvalues in ascending order and the maximum value is 2. A linear graph filter can be demonstrated as a matrix $B = Uh(\Lambda)U^\top\in \mathbbm{R}^{N \times N}$, where $h(\lambda)$ is the frequency response function of $B$. Graph filtering is defined as the multiplication of a graph filter $B$ with a graph signal $X$:

\begin{equation}
    s = Bx,
    \label{high1}
\end{equation}
where $s$ is a filtered graph signal. Signals with rapid changes are carried by high-frequency components, and these signals match the discontinuities and ``opposite attraction" features of heterophilic graphs \citep{li2021beyond}. To preserve the high-frequency components and remove the low-frequency ones in $X$, $h(\lambda)$ should be increasing and nonnegative.

For simplicity, we design our frequency response function as:
\begin{equation}
    h(\lambda)=(\frac{\lambda}{2})^k,
\end{equation}
where $k$ is the filter order. Consequently, the graph filtering $B$ becomes:

\begin{equation}
    B = Uh(\Lambda)U^\top  = (\frac{L}{2})^k.
\end{equation}
By performing graph filtering on $X$, we have the filtered feature matrix:

\begin{equation}
    S = BX = (\frac{L}{2})^kX.
    \label{high}
\end{equation}

Performing such a high-pass filter can encode topology information into features. It also increases the differences between a node and its neighbors, i.e., graph signals are discontinuous, which improves the quality of node features. As a result, it will make the downstream tasks easier in real-world scenarios.

\subsection{Graph Learning}
\begin{figure}[!htbp]
\centering
\includegraphics[width=0.5\textwidth]{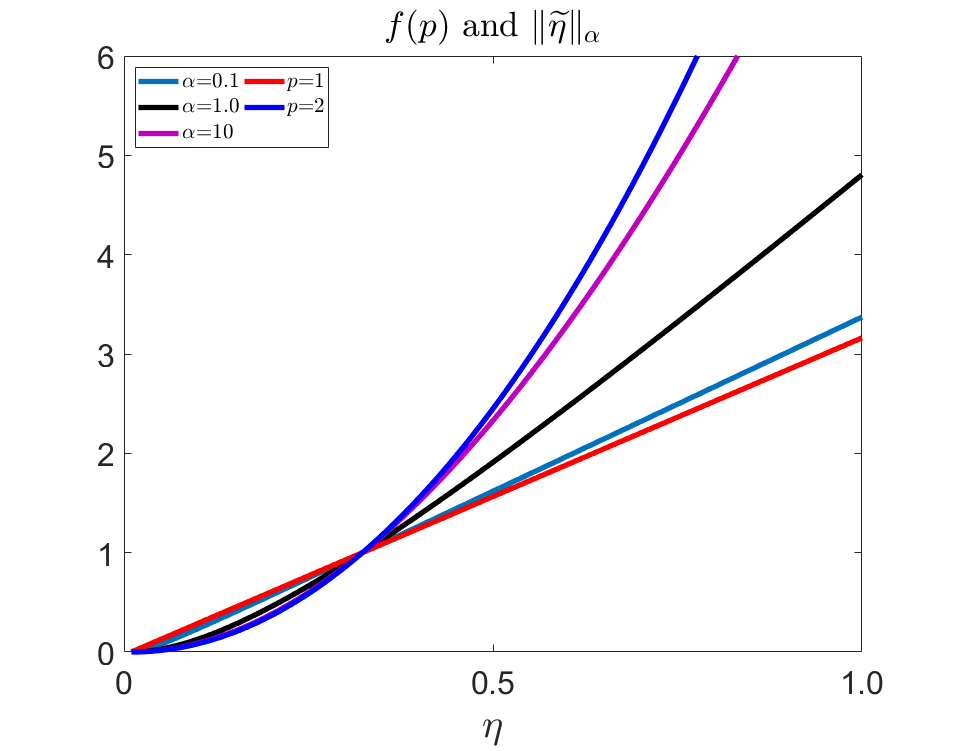} 
\caption{The visualization of $f(p)$ and $\alpha$-norm with different values.}
\label{method}
\end{figure}

 Define $S_{i,\cdot}$ as the $i$-th row of $S$. Dirichlet energy of $S$ measures the smoothness of signals, which is defined as:
\begin{equation}
    \sum_{i,j \in {{E}}_{}}\left\|S_{i,\cdot} - S_{j,\cdot}\right\|_2^2{A}_{ij} = {2\underbrace{Tr(S^\top LS)}_{\text{Dirichlet energy of S}}}.
    \label{dir}
\end{equation}
For practical applications, noise is inevitable. The widely used squared norm used in Eq.(\ref{dir}) is sensitive to large noise. To better characterize the noise of the signals, we could define the function $f(p)$ for the signals $i$ and $j$:

\begin{equation}
    f(p) = \left\|S_{i,\cdot} - S_{j,\cdot}\right\|_2^p,
\end{equation}
where $p$ is a positive integer. Without loss of generality, let's define the noise as $\widetilde{\eta}=\eta \mathbf{1}_{d} \in \mathbbm{R}^{ d}$. In Fig. \Ref{method}, we plot $f(p)$ with $p=1$ and $p=2$. We can see that there is an intersection. When $\eta$ is greater than the intersection, $f(p = 2)$ becomes very sharp while $f(p = 1)$ varies linearly. When $\eta$ is close to 0, $f(p = 1)$ has the maximum slope, which indicates its sensitivity to small variations. Thus, $f(p = 1)$ is robust to large noise and sensitive to small noise, while $f(p = 2)$ has the opposite performance.

To flexibly suit real data, where the magnitude of the noise is often unknown, we define $\alpha$-norm:

\begin{equation}
g(\alpha)=\|S_{i,\cdot}-S_{j,\cdot}\|_\alpha= \frac{(1+\alpha)\left\|S_{i,\cdot}-S_{j,\cdot}\right\|_2^2}{\left\|S_{i,\cdot}-S_{j,\cdot}\right\|_2+\alpha}.
\label{robust_f}
\end{equation}
It is easy to observe that $\alpha$-norm has the following properties:

1) if $\|S_{i,\cdot}-S_{j,\cdot}\|_2 \ll \alpha$, then $\|S_{i,\cdot}-S_{j,\cdot}\|_\alpha \rightarrow \frac{\alpha+1}{\alpha}\left\|S_{i,\cdot}-S_{j,\cdot}\right\|_2^2$.

2) if $\|S_{i,\cdot}-S_{j,\cdot}\|_2 \gg \alpha$, then $\|S_{i,\cdot}-S_{j,\cdot}\|_\alpha \rightarrow (\alpha+1)\left\|S_{i,\cdot}-S_{j,\cdot}\right\|_2$.

3) when $\alpha \rightarrow 0$, $\|S_{i,\cdot}-S_{j,\cdot}\|_\alpha \rightarrow \left\|S_{i,\cdot}-S_{j,\cdot}\right\|_2$, i.e., f(p = 1).

4) when $\alpha \rightarrow \infty$, $\|S_{i,\cdot}-S_{j,\cdot}\|_\alpha \rightarrow \left\|S_{i,\cdot}-S_{j,\cdot}\right\|_2^2$, i.e., f(p = 2).

Therefore, $\alpha$-norm is more general than $f(p)$ and can be adaptive to different levels of noise.
In Fig. \Ref{method}, we plot Eq. (\Ref{robust_f})  with different $\alpha$. It can be seen that $\alpha$-norm is between $f(p = 1)$ and $f(p = 2)$. Besides, $\alpha$-norm has the following characteristics.

\begin{proposition} 
(Shrinking Property) Sensitivity to noise varies monotonically with respect to $\alpha$. It can converge at the extremes, and all values between the upper and lower boundary can be reached.
\label{p1}
\end{proposition} 

\begin{proof}
Consider the case where the noise level is not very low firstly, e.g., $\|S_{i,\cdot}-S_{j,\cdot}\|_2 > 1$. $g(\alpha)$ exists for $\alpha \in (0, \infty)$, thus it's continuous over the domain. Its first order derivative is: $g'(\alpha) = \frac{\left\|S_{i,\cdot}-S_{j,\cdot}\right\|_2^2\left(\left\|S_{i,\cdot}-S_{j,\cdot}\right\|_2-1\right)}{\left(\left\|S_{i,\cdot}-S_{j,\cdot}\right\|_2+\alpha\right)^2}>0.$ Thus $g(\alpha)$ monotonically increases for $\alpha \in (0, \infty)$. The second order derivative is: $g''(\alpha) =\frac{-2\left\|S_{i,\cdot}-S_{j,\cdot}\right\|_2^2\left(\left\|S_{i,\cdot}-S_{j,\cdot}\right\|_2-1\right)}{\left(\left\|S_{i,\cdot}-S_{j,\cdot}\right\|_2+\alpha\right)^3}<0$ and it's closer to 0 when $\alpha$ increases. Thus the change rate of $g(\alpha)$ tends to level off. From properties in 3) and 4), we know that $g(\alpha)$ converges at the upper and lower boundaries and all values between them can be reached. For $\|S_{i,\cdot}-S_{j,\cdot}\|_2 < 1$,  we can reach the same conclusion.
\end{proof}

As a result, introducing $\alpha$ can make the model robust to different levels of noise and flexibly suit different datasets.
Basically, $\alpha$-norm evaluates how smoothly features are distributed across the graph. Minor perturbations to the graph structure might result in an unstable denoising effect. To quantitatively see this, we mathematically prove that $\alpha$-norm has better denoising capability than $f(p=1)$ and $f(p=2)$. 

Assuming the perturbation matrix $\eta^{\prime} \in \mathbbm{R}^{N \times N}$, we define the following graph denoising problem:

\begin{equation}
\min _{A^\prime}\quad\alpha^\prime\left\|A^\prime-\left(A+\eta^{\prime}\right)\right\|_F^2+ (1-\alpha^\prime)\sum_{i, j \in E}\left\|S_{i,\cdot}-S_{j,\cdot}\right\|_2 A^\prime_{i j}
\label{c1}
\end{equation}

\begin{equation}
\min\limits _{A^\prime}\quad\alpha^\prime\left\|A^\prime-\left(A+\eta^{\prime}\right)\right\|_F^2+ (1-\alpha^\prime)\sum_{i, j \in E}\left\|S_{i,\cdot}-S_{j,\cdot}\right\|_2^2 A^\prime_{i j}
\label{c2}
\end{equation}

\begin{equation}
\min\limits _{A^\prime}\quad\alpha^\prime\left\|A^\prime-\left(A+\eta^{\prime}\right)\right\|_F^2+ (1-\alpha^\prime)\sum_{i, j \in E}\left\|S_{i,\cdot}-S_{j,\cdot}\right\|_\alpha A^\prime_{i j},
\label{c3}
\end{equation}
where $A^\prime$ denotes the refined graph from noise data. Eqs. (\ref{c1})-(\ref{c3}) denote the denoising process with $f(p=1)$, $f(p=2)$ and $\alpha$-norm, respectively. 

\begin{proposition}
$\alpha$-norm produces a smoother solution and has better compatibility with noisy nodes than $f(p=1)$ and $f(p=2)$.

\label{p2}
\end{proposition} 

\begin{proof} 
 Setting the gradient of Eqs. (\Ref{c1}-\Ref{c3}) w.r.t. $A^\prime$ to zero, we obtain:
 
\begin{equation}
A^\prime=A+ \frac{2 \alpha^\prime\eta^{\prime}-(1-\alpha^\prime)  \frac{\partial \sum_{i, j \in E}\left\|S_{i,\cdot}-S_{j,\cdot}\right\|_2 A_{i j}^{\prime}}{\partial A^{\prime}}}{2\alpha^\prime}
\end{equation}

\begin{equation}
A^\prime=A+ \frac{2 \alpha^\prime\eta^{\prime}-(1-\alpha^\prime)  \frac{\partial \sum_{i, j \in E}\left\|S_{i,\cdot}-S_{j,\cdot}\right\|_2^2 A_{i j}^{\prime}}{\partial A^{\prime}}}{2\alpha^\prime}
\end{equation}

\begin{equation}
A^\prime=A+ \frac{2 \alpha^\prime\eta^{\prime}-(1-\alpha^\prime)  \frac{\partial \sum_{i, j \in E}\left\|S_{i,\cdot}-S_{j,\cdot}\right\|_\alpha A_{i j}^{\prime}}{\partial A^{\prime}}}{2\alpha^\prime},
\end{equation}

Based on proposition \Ref{p1}, $\frac{\partial \sum_{i, j \in E}\left\|S_{i,\cdot}-S_{j,\cdot}\right\|_\alpha A_{i j}^{\prime}}{\partial A^{\prime}}$ in general has more freedom to be closer to $\frac{2 \alpha^\prime\eta^{\prime}  }{1-\alpha^\prime}$ than $\frac{\partial \sum_{i, j \in E}\left\|S_{i,\cdot}-S_{j,\cdot}\right\|_2 A_{i j}^{\prime}}{\partial A^{\prime}}$ and $\frac{\partial \sum_{i, j \in E}\left\|S_{i,\cdot}-S_{j,\cdot}\right\|_2^2 A_{i j}^{\prime}}{\partial A^{\prime}}$. When $\frac{\partial \sum_{i, j \in E}\left\|S_{i,\cdot}-S_{j,\cdot}\right\|_\alpha A_{i j}^{\prime}}{\partial A^{\prime}}\rightarrow \frac{2 \alpha^\prime\eta^{\prime}  }{1-\alpha^\prime}$, $A^\prime \rightarrow A$. Thus, Eq. (\Ref{c3}) can make the refined graph closer to the initial one, which shows a stronger capability of denoising than the others.

\end{proof}

Motivated by Proposition \Ref{p2}, we can obtain a robust graph $G\in\mathbbm{R}^{N\times N}$ from signals $S$ by a novel cost function:

\begin{equation}
\min _{G}\sum_{i=1}^N \sum_{j=1}^N\left\|S_{i,\cdot} - S_{j,\cdot}\right\|_{\alpha}{G}_{ij}.
\label{fir_tem}
\end{equation}

Since this is an unconstrained problem, we need to design some regularizers for $G$. Inspired by the progress of contrastive learning, we hope to pull positive samples closer while pushing negative samples further away. Mathematically, for sample $x$,

\begin{equation}
G_{x, x^{+}}>>G_{x, x^{-}},
\label{CLP}
\end{equation}
where $x^{+}$ and $x^{-}$ represent positive and negative samples, respectively. Though the connected nodes have different labels in the heterophilic graph, they still share some similarities in most situations, which motivates us to treat edge-connected nodes as positive pairs. Let's define a neighborhood indicator variable:

$$\mathcal{Y}_{ij}=
\begin{cases}
1,& \text{nodes $i$ and $j$ are connected, thus are}\\
&\text{positive samples.}\\
0,& \text{nodes $i$ and $j$ are not connected.}
\end{cases}$$

Afterwards, we design our regularizer as:

\begin{equation}
	\mathcal{J}=\sum_{i=1}^{N} \sum_{j=1}^{N}-\mathcal{Y}_{ij}\log  \frac{\exp \left(G_{i j}\right)}{\sum_{p \neq i}^{N} \exp \left(G_{i p}\right)}.
	\label{contrastY}
\end{equation}

In fact, the graph is sparse and most nodes are not connected. Even if two nodes are not connected in $A$, they may nonetheless have a close bond. This motivates us to update the neighbors at every epoch rather than fixing them by the initial data. We propose to calculate $\mathcal{Y}$ as:

\begin{equation}
        \mathcal{Y}_{ij}=\left\{
        \begin{aligned}
			1,&     & \text { if} \quad  \frac{\left|G_{ij}\right|+\left|G_{ji}\right|}{2}\geq \epsilon. \\
			0,&     &\text { otherwise} .
		\end{aligned}
		\right.
	\label{computnbrs0and1}
\end{equation}
where $\epsilon$ is a threshold number to remove some noisy neighbors. By Eq. (\ref{computnbrs0and1}), the positive samples are updated adaptively during the optimization phase. Overall, our total cost function can be formulated through trade-off parameter $\beta$:

\begin{equation}
\begin{aligned}
	&\min _{G} \sum_{i=1}^N \sum_{j=1}^N\left\|S_{i,\cdot} - S_{j,\cdot}\right\|_{\alpha}{G}_{ij} \\
        &+\beta \sum_{i=1}^{N} \sum_{j=1}^{N}-\mathcal{Y}_{ij}\log \frac{\exp \left(G_{i j}\right)}{\sum_{p \neq i}^N \exp \left(G_{i p}\right)}.\\
	\label{model}
\end{aligned}
\end{equation}

\subsection{Optimization}
The gradient descent method is applied to solve $G$. Its derivative at epoch $t$ is derived as follows:
\begin{equation}
	\nabla_{1}^{(\mathrm{t})}+\beta\nabla_{2}^{(t)}.
	\label{updateg}
\end{equation}
The first term is short for:
\begin{equation}
    \left\|S_{i,\cdot} - S_{j,\cdot}\right\|_{\alpha}.
\end{equation}
Define $M^{(\mathrm{t}-1)}=\sum_{p \neq i}^N \exp \left(G_{i p}^{(t-1)}\right)$, the second term is:
\begin{equation}
	\nabla_{2}^{(t)}=\left\{\begin{array}{l}
		-1+\frac{ \sum_{j^{'}=1}^{N}\mathcal{Y}_{ij^{'}}^{(t-1)}\exp \left(G_{i j}^{(t-1)}\right)}{M^{(t-1)}}, \text { if \enspace} \mathcal{Y}^{(t-1)}_{ij}= 1. \\
		0,\text { otherwise}. 
	\end{array}\right.
\end{equation}

\begin{figure}[!htbp]
\centering
\includegraphics[width=0.3\textwidth]{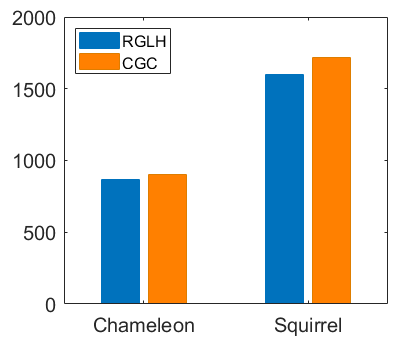} 
\caption{Running time(s) of RGLH and CGC on Chameleom and Squirrel datasets.}
\label{time}
\end{figure}

Then we use Adam optimization to calculate $G$ and $\mathcal{Y}$ alternatively. Besides, we initialize $G$ with the feature's relationship $SS^T$. Our method's computational complexity is $O(N^2)$. Comparatively, the computational complexity of related methods FGC, AGC, and CGC are $O(N^3)$, $O(N^2d)$ and $O(N^2)$, respectively. We compare our running time with CGC in Fig. \ref{time}. It can be seen that our method doesn't cost extra time. Algorithm 1 sums up the whole process.

\begin{algorithm}[htbp]
    \label{algorithm 1}
    \caption{RGSL}
	\label{label}
	\begin{algorithmic}[1]
		\REQUIRE
		adjacency matrix $A$, feature $X$, the order of graph filtering $k$, parameters $\alpha$ and $\beta$, the number of clusters $c$. $D=(A+I)\mathbf{1}_{N}$.
		\ENSURE
		$c$ clusters
		\STATE $L=I-D^{-\frac{1}{2}}(A+I)D^{-\frac{1}{2}}$ 
		\STATE Graph filtering by Eq. (\ref{high})
		\WHILE{convergence condition does not meet}
		\STATE Update $G$ in Eq. (\ref{updateg}) via Adam
		\STATE Update $\mathcal{Y}$ by Eq. (\ref{computnbrs0and1})
		\ENDWHILE
		\STATE $C=\frac{(\left|G\right|+\left|G\right|^{\top})}{2}$
	\end{algorithmic}
\end{algorithm}

\section{Experiments on clustering}\label{experimental1}
We experimentally investigate the performance of RGSL for clustering tasks with real-world benchmark datasets. We perform spectral clustering on the learned graph from RGSL.

\subsection{Datasets}
The used datasets consist of page-page networks on specific topics from Wikipedia called Chameleon and Squirrel \citep{rozemberczki2021multi}; webgraphs gathered by Carnegie Mellon University from computer science departments in several universities\footnote{http://www.cs.cmu.edu/afs/cs.cmu.edu/project/theo-11/www/wwkb/}, including Wisconsin, Cornell, Texas, and Washington; the film-director-actor-writer network's actor-only subgraph called Actor \citep{tang2009social}; Roman-empire is based on Roman Empire article from English Wikipedia, which is selected since it's one of the longest articles on Wikipedia \citep{platonov2023critical}. To compute the homophily of the above graphs, we use the definition given by \citep{Pei2020Geom-GCN}:

\begin{equation}
        homophily = \frac{1}{N}\sum_{v \in\mathcal{V}}{\gamma}_{v} \enspace and \enspace {\gamma}_{v} = \frac{\left|\{u \in \mathcal{N}_{v}|\ell(u)=\ell(v)\}\right|}{\left|\mathcal{N}_{v}\right|},
	\label{homopily}
\end{equation}
where $\mathcal{N}_{v}$ are the neighbors of node $v$ and $\ell$($v$) is the label of node $v$. Additionally, we examine the sparsity of the initial graph $A$, i.e., the percentage of element 1. As seen in Table \ref{tab:dataset}, the presented graph is indeed quite sparse, which encourages us to learn a rich graph.

\subsection{Baseline methods}
To our knowledge, no method is focusing on heterophilic graph clustering. We compare RGSL with several popular clustering methods, including DAEGC \citep{wang2019attributed}, MSGA \citep{wang2021multi}, FGC \citep{FGC}, DCRN \citep{liu2022deep}, VGAE-based methods \citep{zhu2022collaborative} and CGC \citep{CGC}. DAEGC \citep{wang2019attributed} proposes a graph attentional autoencoder to aggregate neighborhood information with different weights. MSGA \citep{wang2021multi} proposes a GCN-based feature extraction method to achieve representations with self-attention mechanisms and a self-supervised module to supervise the representation learning. FGC \citep{FGC} proposes a graph learning method with high-order information for clustering. DCRN \citep{liu2022deep} is a feature-level contrastive clustering method with a siamese network. \citep{zhu2022collaborative} designs a successful generation technique leveraging pseudo-labels and boosts the performance of VGAE-based methods. CGC \citep{CGC} is the only shallow method that considers heterophily for clustering. To further support our claim, we additionally replace the $\alpha$-norm  with $f(p=2)$ and mark this model as RGSL-.

\begin{table}[htbp]
\centering

\renewcommand{\arraystretch}{1.}
\caption{The statistics of datasets.}
\resizebox{.8\textwidth}{!}{
\begin{tabular}{lrrrrrr}
    \toprule
    Dataset    & Nodes & Edges & Features  & Classes  &Homophily &Sparsity\\
    \midrule
    Chameleon &2277&31371&2325&5&0.25&0.61\%\\
    Squirrel  &5201 &198353 & 2,089 & 5 & 0.22 &0.73\%\\
    Actor  &7600 &33544 & 931 & 5 & 0.22 &0.12\%\\
    Wisconsin &251&515&1703&5&0.15&1.63\%\\
    Cornell   &183&298&1703&5&0.11&1.78\%\\
    Washington   &217&446&1703&5&0.27&1.89\%\\
    Texas     &183&325&1703&5&0.06&1.94\%\\
    Roman-empire     &22662&32927&300&18&0.05&0.01\%\\
    

    \bottomrule
    \hline
\end{tabular}}
\label{tab:dataset}
\end{table}



\subsection{Setup}
Learning rate is set to 0.1 on Wisconsin and 0.01 on other datasets. Threshold $\epsilon$ is set to 1 on Wisconsin and 0.001 on other datasets. Other parameter settings ($k$, $\alpha$, $\beta$) are: (5, 5, 1) on Chameleon; (2, 5, 100) on Squirrel; (1, 50, 100) on Wisconsin; (17, 0.001, 100) on Cornell; (4, 0.01, 0.001) on Texas; (8, 0.01, 100) on Washington. (5, 0.001, 1) on Roman-empire. We use three widely used metrics to evaluate the effectiveness of clustering: (1) Clustering accuracy (ACC), which assesses the performance of label matching clustering results; (2) Normalized Mutual Information (NMI) quantifies the mutual information entropy between cluster labels and the ground truth labels. The Macro F1-score (F1) is the harmonic mean of precision and recall. We search for the best parameters for each method. As for baselines, most results are directly quoted from CGC \citep{CGC} and \citep{zhu2022collaborative}, while the rest results follow original papaer's suggestions to perform grid search.

\subsection{Results of clustering}
Table \Ref{1} reports the clustering results on heterophilic graphs. RGSL significantly dominates the baselines DAEGC and MSGA, which are based on graph autoencoder. Compared to FGC, our method consistently outperforms it. This is because FGC uses low-pass filter, which is incompatiable with the property of heterophilic graph. Our method also beats DCRN in general, which is based on contrastive learning. Moreover, our method  improves the results of recent methods: ARVGA-Col-M, RWR-Col-M, GMM-Col-M.
These results prove the advantages of our method over existing graph clustering methods. Though it is not using deep neural networks, our method still achieves promising performance. Compared to the most recent clustering method CGC, RGSL can still surpass it on all cases, especially on Cornell, Texas and Roman-empire, which are the datasets with high heterophily. Thus, our robust model is more applicable in real world. Furthermore, RGSL surpasses RGSL- in most cases, which validates the benefit of $\alpha$-norm.

 \setlength{\tabcolsep}{0.5pt}
\begin{table*}[t]\scriptsize
 \centering
 \renewcommand{\arraystretch}{1.}
 \caption{Clustering performance on heterophilic graph datasets.}
\resizebox{1.\textwidth}{!}{
  \begin{tabular}{l rcl | rcl| rcl| rcl| rcl| rcl| rcl}
   \toprule
   Methods\,\,   &\multicolumn{3}{c}{{Chameleon}} &\multicolumn{3}{c}{{Squirrel}} & \multicolumn{3}{c}{{Wisconsin}}  &\multicolumn{3}{c}{{Cornell}} & \multicolumn{3}{c}{{Texas}} & \multicolumn{3}{c}{{Washington}} &\multicolumn{3}{c}{{Roman-empire}}\\
   \cmidrule(lr){2-4} \cmidrule(lr){5-7} \cmidrule(lr){8-10} \cmidrule(lr){11-13}\cmidrule(lr){14-16}\cmidrule(lr){17-19}\cmidrule(lr){20-22}
    & {ACC\%} & {NMI\%} & {F1\%}  & {ACC\%} & {NMI\%} & {F1\%}  & {ACC\%} & {NMI\%} & {F1\%}& {ACC\%} & {NMI\%} & {F1\%}& {ACC\%} & {NMI\%} & {F1\%}& {ACC\%} & {NMI\%} & {F1\%}& {ACC\%} & {NMI\%} & {F1\%}\\
      \midrule

   DAEGC \citep{wang2019attributed} &32.06&6.45&5.46&25.55&2.36&\textbf{24.07}& 39.62 & 12.02 & 6.22 & 42.56 & 12.37 & 30.20  & 45.99 & 11.25 & 18.09 &46.96&17.03&14.54  &21.23 &12.67 &5.64\\
    MSGA \citep{wang2021multi} 
    &27.98&6.21&9.85&27.42 &4.31&18.41  &54.72&16.28&12.03 &50.77&14.05&11.05 &57.22&12.13&15.91&49.81&6.38&5.34&19.31 &12.25 &5.22\\


   ARVGA-Col-M \citep{zhu2022collaborative} &-&-&-&-&-&-&54.34&11.41&-&-&-&-&59.89&16.37&-&60.43&15.58&-&- &- &-\\
   RWR-Col-M \citep{zhu2022collaborative}&-&-&-&-&-&-&53.58&16.25&-&-&-&-&57.22&13.82&-&63.48&22.05&-&- &- &-\\
   GMM-Col-M \citep{zhu2022collaborative}&-&-&- &-&-&- &51.70 &9.68&- &-&-&- &58.29&11.55&- &60.86&20.56&-&- &- &-\\

   FGC \citep{FGC} &36.50&11.25&36.80&25.11&1.32&22.13 & 50.19 & 12.92 & 25.93 & 44.10 & 8.60 & 32.68 & 53.48 & 5.16 & 17.04&57.39&21.38&35.37&14.46 &4.86 &11.60
   \\
   
   DCRN \citep{liu2022deep} &34.52&9.11&26.92&30.69&6.84&24.00 &\textbf{57.74}&19.86&37.63 &51.32&9.05&26.73 &63.10&24.14&30.64&55.65&14.15&26.32&32.57 &29.50 &17.09\\
   CGC \citep{CGC}&36.43 &11.59 &32.93 &27.23 &2.98 &20.57  &55.85 &23.03 &27.29 &44.62 &14.11 &21.91 &61.50 &21.48 &27.20 &63.20&25.94&30.75 &30.16 &27.25 &17.22\\
   
   
    \midrule
   RGSL- &37.59&11.36&36.90&29.67&7.67&17.89&56.23&16.17&34.28 &53.33&\textbf{29.38}&47.45 &65.24&29.25&39.02&61.30&24.25&39.43&32.58 &28.69 &\textbf{18.83}\\
    
   RGSL   &\textbf{38.52}&\textbf{12.79}&\textbf{37.54}&\textbf{30.74}&\textbf{8.74}&18.57& 56.60 & \textbf{28.57} & \textbf{43.08} & \textbf{57.44} & 28.95 & \textbf{48.49}  &\textbf{72.19} &\textbf{37.86} &\textbf{40.24} &\textbf{66.09}&\textbf{29.79}&\textbf{39.66}&\textbf{34.57} &\textbf{31.23} &18.66 \\

   \bottomrule
  \end{tabular}}
  \label{1}
\end{table*}

\subsection{Ablation Study}
To see the the impact of high-pass filter, we test our method without it and mark this model as "RGSL w/o $\mathcal{F}$". The results are illustrated in Table \Ref{ablation}. We can observe degradation in general, thus high-pass filter is helpful in enhancing the attribute quality. To test the regularizer's effect, we replace it with the popular Frobenius norm:
\begin{equation}
\begin{aligned}
&\min _{G}\sum_{i=1}^N \sum_{j=1}^N\left\|S_{i,\cdot} - S_{j,\cdot}\right\|_{\alpha}{G}_{ij} + \beta\left \|G\right \|_{F}^{2} \\
 &\text { s.t. }  G_{\mathrm{ij}} \geq 0, G_{i,\cdot} \mathbf{1}_{N}=1,
\label{23}
\end{aligned}
\end{equation}
which is marked as ``RGSL w/o $\mathcal{J}$". According to Table \Ref{ablation}, the performance is decreased in most cases. It indicates that contrastive regularizer indeed improves the graph quality. To further verify the benefit of $\mathcal{Y}$, we just remove it and select positive samples from neighbors $\mathcal{N}_{i}$ by $K$-nearest neighbors ($K=10$), which is marked as "RGSL w/ $\mathcal{N}$". 
Then the total loss becomes:

\begin{equation}
\begin{aligned}
&\min _{G}\sum_{i=1}^N \sum_{j=1}^N\left\|S_{i,\cdot} - S_{j,\cdot}\right\|_{\alpha}{G}_{ij}\\
&+\beta \sum_{i=1}^{N} \sum_{j \in \mathcal{N}_{i}}-\log \frac{\exp \left(G_{i j}\right)}{\sum_{p \neq i}^N \exp \left(G_{i p}\right)}.
\label{modelCGC}
\end{aligned}
\end{equation}
Similarly, we have a performance drop. Thus, adaptive positive samples are helpful. Furthermore, we replace $\mathcal{Y}$ with $A$ and denote the model as ``RGSL w/ $A$". Compared with RGSL, the drop in performance is obvious in most cases. This also suggests that the learned graph is more accurate than the original data in characterizing the relations among nodes.

\begin{table*}[t]
 \centering
 \renewcommand{\arraystretch}{1.}
 \caption{Results of ablation study.}
 \resizebox{1.\textwidth}{!}{
  \begin{tabular}{l rcl| rcl| rcl| rcl| rcl| rcl| rcl}
   \toprule
   Methods & \multicolumn{3}{c}{{Chameleon}} & \multicolumn{3}{c}{{Squirrel}} & \multicolumn{3}{c}{{Wisconsin}} & \multicolumn{3}{c}{{Cornell}}& \multicolumn{3}{c}{{Texas}}& \multicolumn{3}{c}{{Washington}} &\multicolumn{3}{c}{{Roman-empire}}\\
   \cmidrule(lr){2-4} \cmidrule(lr){5-7} \cmidrule(lr){8-10} \cmidrule(lr){11-13}\cmidrule(lr){14-16}\cmidrule(lr){17-19}\cmidrule(lr){20-22}
    & {ACC\%} & {NMI\%} & {F1\%}  & {ACC\%} & {NMI\%} & {F1\%}  & {ACC\%} & {NMI\%} & {F1\%} & {ACC\%} & {NMI\%} & {F1\%}& {ACC\%} & {NMI\%} & {F1\%}& {ACC\%} & {NMI\%} & {F1\%}& {ACC\%} & {NMI\%} & {F1\%}\\
   \midrule
   
  RGSL w/o $\mathcal{F}$ &30.30&10.77&19.28&21.94&1.53&21.93  &52.45&10.53&22.48  &51.28&26.05&43.94    &54.55&4.99&19.41&62.17&25.01&36.87&32.83&30.86&17.90\\

  RGSL w/o $\mathcal{J}$ &32.54&13.94&29.58&26.88&4.47&19.87 &53.58&20.21&26.65&53.85 & \textbf{30.71}   & 47.24 &68.45&33.02&47.97&63.91&27.23&38.62&33.09&30.22&17.90\\

    RGSL w/ $\mathcal{N}$ &30.78&16.45&25.30&24.57&1.16&\textbf{24.12} &55.85&15.51&33.45&51.28 & 26.97  & \textbf{49.53} &65.77&31.31&44.11 &61.74&26.03&\textbf{40.20}&33.45&29.83&\textbf{18.82}\\

    RGSL w/ $A$ &35.00&\textbf{17.89}&30.22&27.46&5.18&19.70 &54.72&13.98&33.33&54.36 & 24.25   & 40.12 &68.98&30.37&\textbf{48.92}&62.18&23.53&35.84&33.33&30.49&18.17\\


   \midrule

   RGSL   &\textbf{38.52}&12.79&\textbf{37.54}&\textbf{30.74}&\textbf{8.74}&18.57& \textbf{56.60} & \textbf{28.57} & \textbf{43.08} & \textbf{57.44} & 28.95 & 48.49  &\textbf{72.19} &\textbf{37.86} &40.24 &\textbf{66.09}&\textbf{29.79}&39.66 &\textbf{34.57} &\textbf{31.23} &18.66\\

   \bottomrule
  \end{tabular}}
  \label{ablation}
  
\end{table*}

\subsection{Parameter Analysis}
\begin{figure*}[!htbp]
\centering
\includegraphics[width=1.0\textwidth]{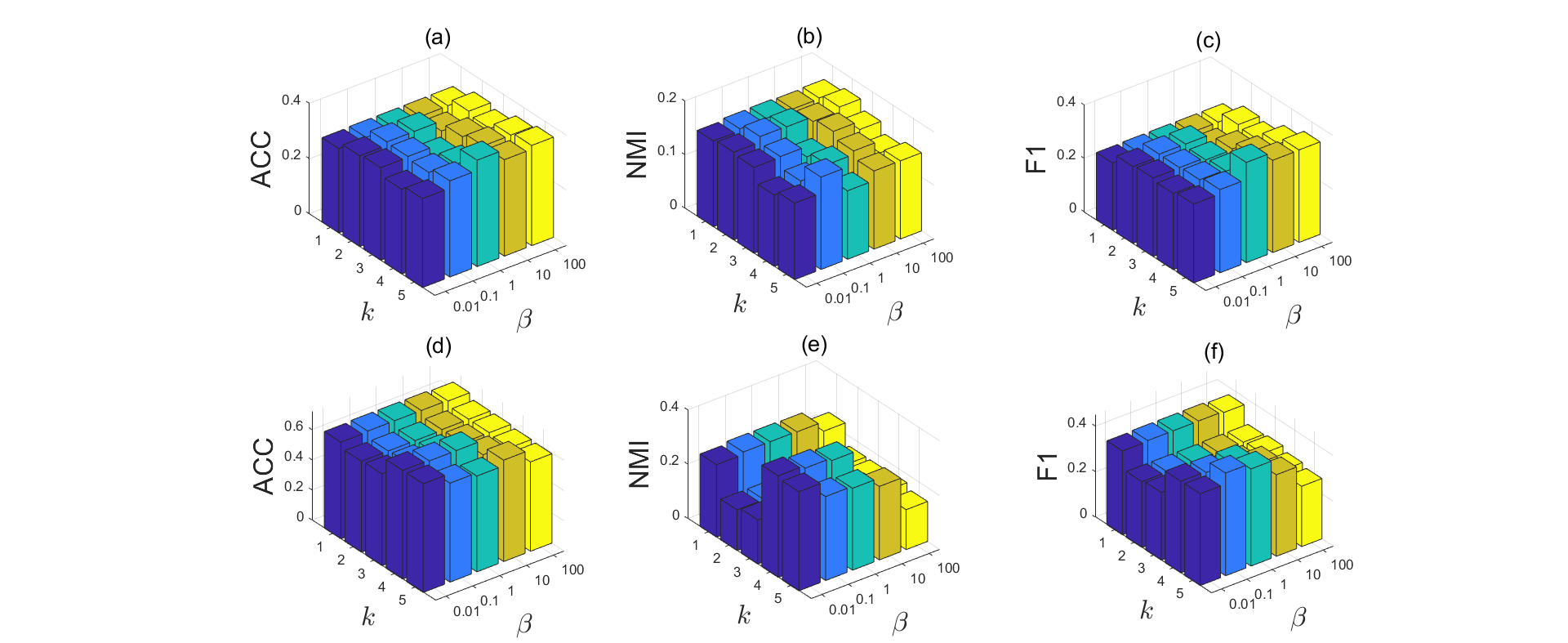} 
\caption{Sensitivity analysis of $k$ and $\beta$ on Chameleon (a-c) and Texas (d-f).}
\label{kbeta}
\end{figure*}

There are three parameters in our method, including filter order $k$, trade-off parameters $\alpha$, and $\beta$. We set different $k$ and $\beta$ values on Chameleon and Texas to see their influence. The results are illustrated in Fig. \Ref{kbeta}. It shows that our method can work well for a large range of parameters. In particular, we don't need to perform filtering too many times.

\begin{figure*}[!htbp]
\centering
\includegraphics[width=1\textwidth]{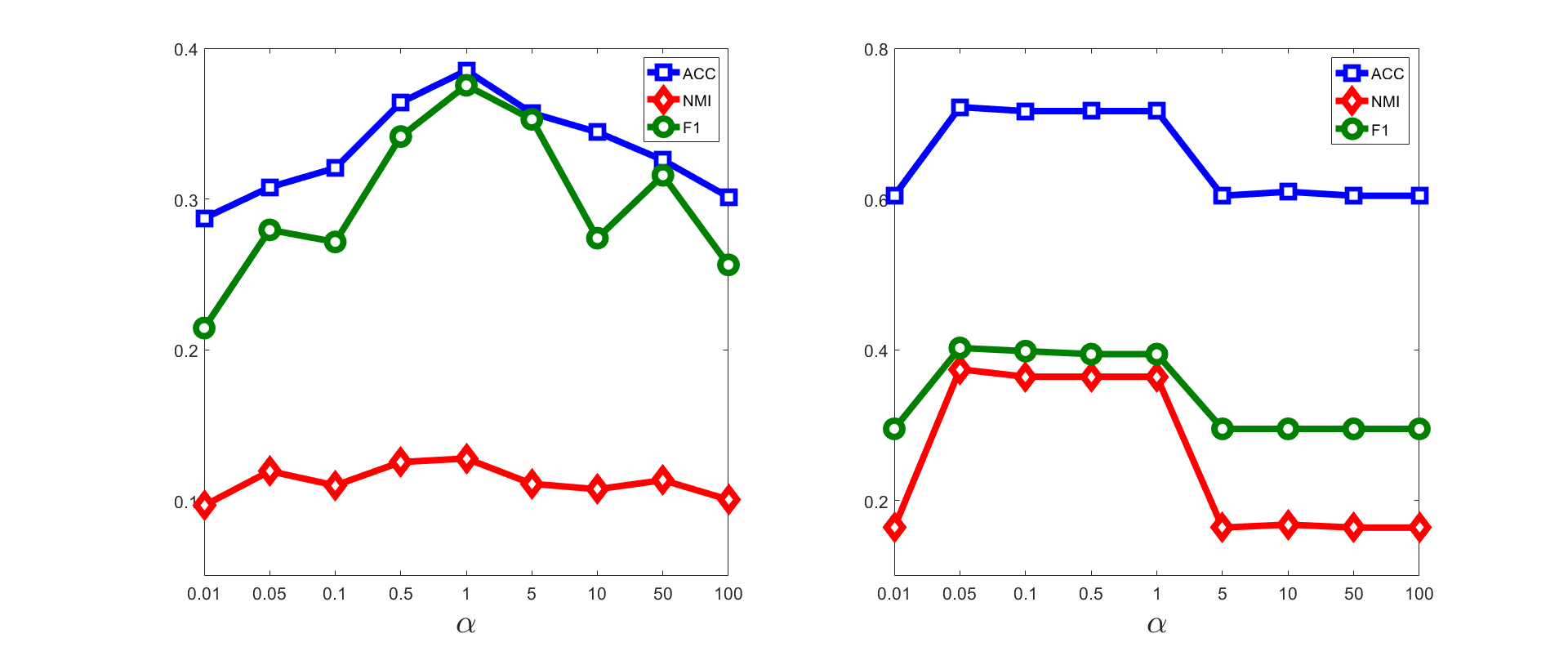} 
\caption{Sensitivity analysis of $\alpha$ on Chameleon (left) and Texas (right).}
\label{alpha}
\end{figure*}

$\alpha$ is searched in grid of [0.01, 0.05, 0.1, 0.5, 1, 5, 10, 50, 100] and Fig. \Ref{alpha} shows how it affects clustering results. We can observe that our model is sensitive to it and too large or small $\alpha$ may cause poor results. This indicates that the noise level is crucial and $\alpha$-norm is a good way to handle this issue. According to our experience, small $\alpha$ is always set on low homophlily graph datasets since they are more possible to have large noise. For instance, Texas and Cornell report the lowest and the second lowest homophily score, respectively. $\alpha$ is set to 0.01 on them. In fact, our graph learning approach will improve graph homophily, which is responsible for clustering accuracy. For example, the $C$'s homophily is 0.37 on Texas and 0.27 on Cornell, which are 613 and 240 times big w.r.t. raw graph, respectively. By contrast, datasets with relatively high homophlily always have a large $\alpha$. Chameleon has a high homophlily score and $\alpha$ is set to 5 on it. 

Besides, without knowing homophily, $\alpha$ can also be intuitively selected by Dirichlet energy. Outliers are the nodes either with only one edge or too many edges, which are more possible to be noisy than other nodes. Large noise's Dirichlet energy is larger than the small one. Thus the ratio of the Dirichlet energy of them to the whole graph can be used as a measure. We take the average value for each outlier. The ratio (and outlier numbers) are 0.139 (88) on Texas, 0.194 (113) on Cornell, and 0.017 (557) on Chameleon, which share the same trend as before.






\section{Experiments on node classification}\label{experimental2}
In this section, we examine the performance of our graph structure learning approach on semi-supervised node classification task.

After obtaining the graph learned from RGSL, we input it into the classical local and global consistency (LGC) \citep{zhou2004learning} technique to carry out semi-supervised classification task. Specifically, LGC obtains the classification function ${M}\in{R}^{N \times g}$ by solving the following problem:
\begin{equation}
\min _M \operatorname{Tr}\left\{M^T L' M+\gamma(M-C)^\top(M-C)\right\},
\end{equation}
where $L'$ denotes the the graph Laplacian matrix computed using inputted $G$, $g$ denotes the number of classes and ${C}\in{R}^{N \times g}$ denotes label matrix, in which $c_{ij}=1$ if the $i$-th node belongs to the $j$-th class, otherwise $c_{ij}=0$. 

\begin{figure*}[!htbp]
\centering
\includegraphics[width=1.0\textwidth]{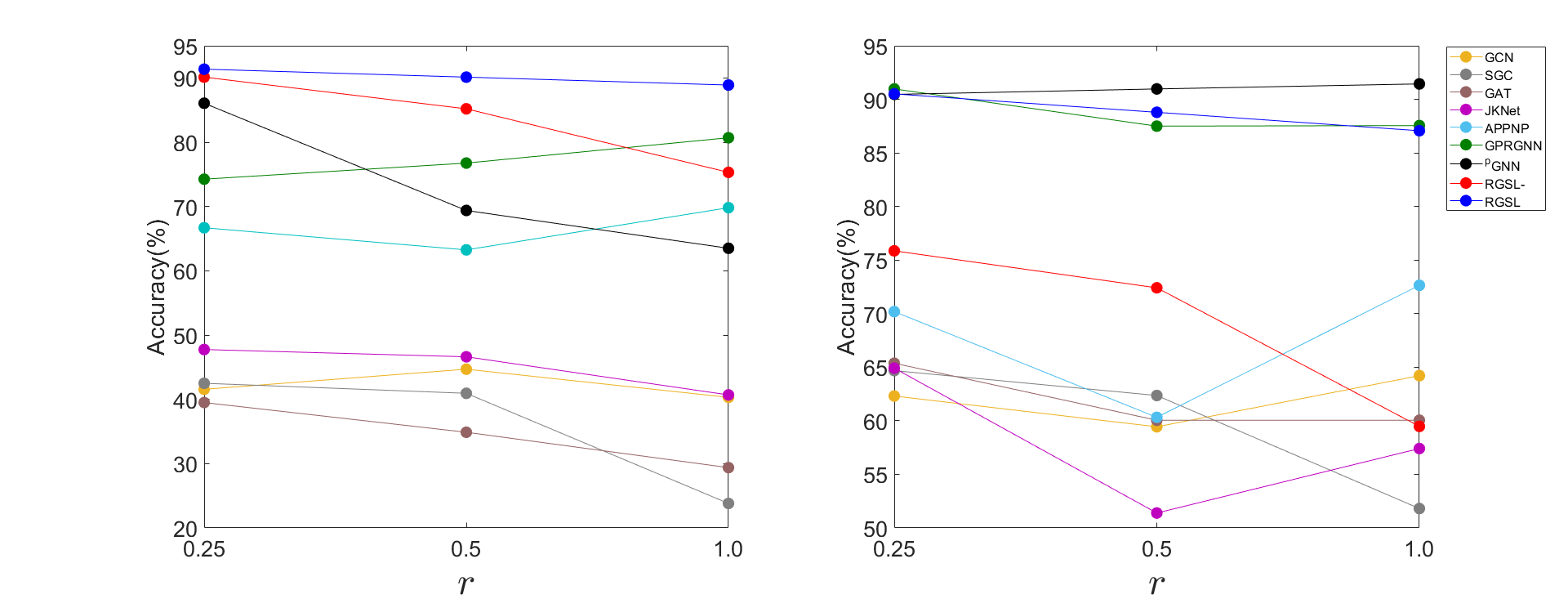} 
\caption{Accuracy (\%) on graphs with noisy edges on Texas (left) and Wisconsin (right).}
\label{noise}
\end{figure*}

\subsection{Baseline methods}
We compare RGSL with seven state-of-the-art algorithms for semi-supervised node classification. They are GCN \citep{kipf2016semi}, SGC \citep{wu2019simplifying}, GAT \citep{velivckovic2018graph}, JKNet \citep{xu2018representation}, APPNP \citep{gasteiger2018predict}, GPRGNN \citep{chien2021adaptive}, $^{p}$GNN \citep{fu2022p}. GCN \citep{kipf2016semi} and SGC \citep{wu2019simplifying} are popular spectral-based GNN models. GCN stacks multiple layers and emploies a nonlinear activation function after each layer, while SGC simplifies the propagation with a linear model. GAT \citep{velivckovic2018graph} proposes masked self-attentional layers to choose important neighbors. JKNet \citep{xu2018representation} is flexible to use different neighborhood ranges for each node. APPNP \citep{gasteiger2018predict} combines PageRank and GCN to pass information. GPRGNN \citep{chien2021adaptive} assigns a learnable weight to each stage of the feature propagation process and achieves significant improvements on heterophilic graphs. $^{p}$GNN \citep{fu2022p} proposes a new massage-passing architecture based on graph divergence to deal with noise and heterophily, and it's the most recent method.

\subsection{Setup}
For an unbiased comparison, the node classification task is conducted with the same settings in $^{p}$GNN \citep{fu2022p} and some results are directly quoted from it. Specifically, a dense splitting approach (60\%, 20\%, 20\%) is used to randomly divide the graphs into training, validation, and testing sets. Note that, different from \citep{chien2021adaptive}, Chameleon and Squirrel are not processed to be undirected graphs. Hyperparameters settings for other methods are: it consists of 2 layers, with a maximum of 1000 epochs, an early stopping threshold set at 200, and a weight decay of either 0 or 0.0005. Other hyperparameters (Number of hidden units, Learning rate, Dropout rate, $K$) are: (16, 0.001/0.01/0.05, 0/0.5, 4/6/8) and $\mu$ = 0.01/0.1/0.2/1/10 on $^{p}$GNN; (16, 0.001/0.01, 0/0.5, None) on GCN; (16, 0.2/0.01, 0/0.5, 2) on SGC; (8, 0.001/0.005, 0/0.6, None) and number of attention heads is 8 on GAT; (16, 0.001/0.01, 0/0.5, 10), $\alpha$ = 0.1/0.5/0.7/1, number of GCN based layers is 2 and the layer aggregation is LSTM with 16 channels and 4 layers on JKNet; (16, 0.001/0.01, 0/0.5, 10) and $\alpha$ = 0.1/0.5/0.7/1 on APPNP; (16, 0.001/0.01/0.05, 0/0.5, 10), $\alpha$ = 0/0.1/0.2/0.5/0.7/0.9/1 and dprate = 0/0.5/0.7 on GPRGNN.


\subsection{Results of classification}
Table \ref{semi} summarizes the classification results. It shows that RGSL outperforms many deep learning methods substantially and it achieves the best performance on most datasets. We can also observe that classical GNNs produce inferior results due to their inherent homophily assumption. Both GPRGNN and $^{p}$GNN pay attention to heterophilic issue and achieve better accuracy than other GNNs-based methods. In particular, $^{p}$GNN produces the second best performance since it can learn aggregation weights in an adaptive manner and is robust to noisy edges. Though these endeavors are based on neural networks, our simple approach still outperforms them.
 \setlength{\tabcolsep}{1pt}\begin{table}
 \centering
 \renewcommand{\arraystretch}{1.}
 \caption{Semi-supervised node classification accuracy. The best performance is highlighted in \textbf{bold} and the second one is colored in \textcolor{blue}{blue}.}
  \label{tab:results 1}
  \resizebox{1.\textwidth}{!}{
  \begin{tabular}{l c c c c c c}
   \toprule
   Methods & \multicolumn{1}{c}{{Chameleon}} & \multicolumn{1}{c}{{Squirrel}} & \multicolumn{1}{c}{{Actor}} & \multicolumn{1}{c}{{Wisconsin}}& \multicolumn{1}{c}{{ Texas}} & \multicolumn{1}{c}{{Cornell}}\\

   \midrule

   GCN \citep{kipf2016semi}& 34.54& 25.28 & 31.28 & 61.93 & 56.54 & 51.36\\
   SGC \citep{wu2019simplifying}& 34.76& 25.49 & 30.98 & 66.94 & 59.99 & 44.39\\
   GAT \citep{velivckovic2018graph}& 45.16& 31.41 & 34.11 & 65.64 & 56.41 & 43.94\\
   JKNet \citep{xu2018representation}& 33.28 & 25.82 & 29.77 & 61.08 & 59.65 & 55.34\\
   APPNP \citep{gasteiger2018predict}& 36.18 & 26.85 & 31.26 & 64.59 & 82.90 & 66.47\\
   GPRGNN \citep{chien2021adaptive}  &43.67 & 31.27 & 36.63 & 88.54 & 80.74 & \textcolor{blue}{78.95}\\
   $^{p}$GNN \citep{fu2022p} & \textcolor{blue}{48.77} & \textcolor{blue}{33.60} & \textbf{40.07} & \textcolor{blue}{91.15} & \textcolor{blue}{87.96} & 72.04\\
   
   \midrule

   RGSL & \textbf{49.02} & \textbf{34.97} & \textcolor{blue}{39.78} & \textbf{91.34} & \textbf{91.36} & \textbf{80.60}\\

   \bottomrule
  \end{tabular}}
  \label{semi}
\end{table}

\subsection{Robustness test}
To test how well RGSL performs on graph with noisy edges, we randomly add edges to the graphs and randomly remove the same number of original edges. Define random edge rate $r = \frac{\text{\#random edges}}{\text{\#all edges}}$. The experiments are conducted on Texas and Wisconsin with $r$ = 0.25, 0.5, 1 following \citep{fu2022p}. Fig. \Ref{noise} reports the results with noisy edges. For Texas, we can see that RGSL can get the best performance in all cases and is more stable than other GNN-based methods.
The accuracy of other deep methods change dramatically  with increasing noise,
which suggests that a robust graph is of great
significance when training GNNs in real scenarios. In particular, RGSL- shows a sharp decline when $r$ increases, which validates
our previous claim that $f(p=2)$ is sensitive to large noise. For Wisconsin, stable performance is also given by RGSL. RGSL- performs
extremely poor and is considerably influenced by the noise, which further verifies the necessity of our $\alpha$-norm.

\section{Conclusion} \label{conclude}
In this work, we make the first attempt to learn a robust graph from heterophilic data. To be consistent with the characteristic of heterophilic graph, we design a high-pass filter, which extracts valuable high-frequency information. After filtering, we propose a graph structure learning method to flexibly characterize different levels of noise. An augmentation-free contrastive regularizer is also applied to further refine graph structure with adaptive positive samples. Comprehensive experiments on clustering and semi-supervised learning tasks demonstrate that our approach achieves clear improvements over state-of-the-art methods. It indicates that our simple method can have better performance than the deep neural networks approaches in some real-world cases, which is appealing in practice.

\bibliographystyle{cas-model2-names}

\bibliography{1}

\begin{thebibliography}{49}
\expandafter\ifx\csname natexlab\endcsname\relax\def\natexlab#1{#1}\fi
\providecommand{\url}[1]{\texttt{#1}}
\providecommand{\href}[2]{#2}
\providecommand{\path}[1]{#1}
\providecommand{\DOIprefix}{doi:}
\providecommand{\ArXivprefix}{arXiv:}
\providecommand{\URLprefix}{URL: }
\providecommand{\Pubmedprefix}{pmid:}
\providecommand{\doi}[1]{\href{http://dx.doi.org/#1}{\path{#1}}}
\providecommand{\Pubmed}[1]{\href{pmid:#1}{\path{#1}}}
\providecommand{\bibinfo}[2]{#2}
\ifx\xfnm\relax \def\xfnm[#1]{\unskip,\space#1}\fi
\bibitem[{Chen et~al.(2020a)Chen, Kornblith, Norouzi and Hinton}]{chen2020simple}
\bibinfo{author}{Chen, T.}, \bibinfo{author}{Kornblith, S.}, \bibinfo{author}{Norouzi, M.}, \bibinfo{author}{Hinton, G.}, \bibinfo{year}{2020}a.
\newblock \bibinfo{title}{A simple framework for contrastive learning of visual representations}, in: \bibinfo{booktitle}{International conference on machine learning}, pp. \bibinfo{pages}{1597--1607}.
\bibitem[{Chen et~al.(2020b)Chen, Wu and Zaki}]{chen2020iterative}
\bibinfo{author}{Chen, Y.}, \bibinfo{author}{Wu, L.}, \bibinfo{author}{Zaki, M.}, \bibinfo{year}{2020}b.
\newblock \bibinfo{title}{Iterative deep graph learning for graph neural networks: Better and robust node embeddings}.
\newblock \bibinfo{journal}{Advances in neural information processing systems} \bibinfo{volume}{33}, \bibinfo{pages}{19314--19326}.
\bibitem[{Chien et~al.(2020)Chien, Peng, Li and Milenkovic}]{chien2021adaptive}
\bibinfo{author}{Chien, E.}, \bibinfo{author}{Peng, J.}, \bibinfo{author}{Li, P.}, \bibinfo{author}{Milenkovic, O.}, \bibinfo{year}{2020}.
\newblock \bibinfo{title}{Adaptive universal generalized pagerank graph neural network}, in: \bibinfo{booktitle}{International Conference on Learning Representations}.
\bibitem[{Dwibedi et~al.(2021)Dwibedi, Aytar, Tompson, Sermanet and Zisserman}]{dwibedi2021little}
\bibinfo{author}{Dwibedi, D.}, \bibinfo{author}{Aytar, Y.}, \bibinfo{author}{Tompson, J.}, \bibinfo{author}{Sermanet, P.}, \bibinfo{author}{Zisserman, A.}, \bibinfo{year}{2021}.
\newblock \bibinfo{title}{With a little help from my friends: Nearest-neighbor contrastive learning of visual representations}, in: \bibinfo{booktitle}{Proceedings of the IEEE/CVF International Conference on Computer Vision}, pp. \bibinfo{pages}{9588--9597}.
\bibitem[{Fatemi et~al.(2021)Fatemi, El~Asri and Kazemi}]{fatemi2021slaps}
\bibinfo{author}{Fatemi, B.}, \bibinfo{author}{El~Asri, L.}, \bibinfo{author}{Kazemi, S.M.}, \bibinfo{year}{2021}.
\newblock \bibinfo{title}{Slaps: Self-supervision improves structure learning for graph neural networks}.
\newblock \bibinfo{journal}{Advances in Neural Information Processing Systems} \bibinfo{volume}{34}, \bibinfo{pages}{22667--22681}.
\bibitem[{Fu et~al.(2022)Fu, Zhao and Bian}]{fu2022p}
\bibinfo{author}{Fu, G.}, \bibinfo{author}{Zhao, P.}, \bibinfo{author}{Bian, Y.}, \bibinfo{year}{2022}.
\newblock \bibinfo{title}{$ p $-laplacian based graph neural networks}, in: \bibinfo{booktitle}{International Conference on Machine Learning}, \bibinfo{organization}{PMLR}. pp. \bibinfo{pages}{6878--6917}.
\bibitem[{Gasteiger et~al.(2018)Gasteiger, Bojchevski and G{\"u}nnemann}]{gasteiger2018predict}
\bibinfo{author}{Gasteiger, J.}, \bibinfo{author}{Bojchevski, A.}, \bibinfo{author}{G{\"u}nnemann, S.}, \bibinfo{year}{2018}.
\newblock \bibinfo{title}{Predict then propagate: Graph neural networks meet personalized pagerank}, in: \bibinfo{booktitle}{International Conference on Learning Representations}.
\bibitem[{Jin et~al.(2020)Jin, Ma, Liu, Tang, Wang and Tang}]{jin2020graph}
\bibinfo{author}{Jin, W.}, \bibinfo{author}{Ma, Y.}, \bibinfo{author}{Liu, X.}, \bibinfo{author}{Tang, X.}, \bibinfo{author}{Wang, S.}, \bibinfo{author}{Tang, J.}, \bibinfo{year}{2020}.
\newblock \bibinfo{title}{Graph structure learning for robust graph neural networks}, in: \bibinfo{booktitle}{Proceedings of the 26th ACM SIGKDD international conference on knowledge discovery \& data mining}, pp. \bibinfo{pages}{66--74}.
\bibitem[{Kang et~al.(2022)Kang, Liu, Pan and Tian}]{FGC}
\bibinfo{author}{Kang, Z.}, \bibinfo{author}{Liu, Z.}, \bibinfo{author}{Pan, S.}, \bibinfo{author}{Tian, L.}, \bibinfo{year}{2022}.
\newblock \bibinfo{title}{Fine-grained attributed graph clustering}, in: \bibinfo{booktitle}{Proceedings of the 2022 SIAM International Conference on Data Mining (SDM)}, \bibinfo{organization}{SIAM}. pp. \bibinfo{pages}{370--378}.
\bibitem[{Kipf and Welling(2017)}]{kipf2016semi}
\bibinfo{author}{Kipf, T.N.}, \bibinfo{author}{Welling, M.}, \bibinfo{year}{2017}.
\newblock \bibinfo{title}{Semi-supervised classification with graph convolutional networks}, in: \bibinfo{booktitle}{ICLR}.
\bibitem[{Kong et~al.(2019)Kong, d'Autume, Ling, Yu, Dai and Yogatama}]{kong2019mutual}
\bibinfo{author}{Kong, L.}, \bibinfo{author}{d'Autume, C.d.M.}, \bibinfo{author}{Ling, W.}, \bibinfo{author}{Yu, L.}, \bibinfo{author}{Dai, Z.}, \bibinfo{author}{Yogatama, D.}, \bibinfo{year}{2019}.
\newblock \bibinfo{title}{A mutual information maximization perspective of language representation learning}.
\newblock \bibinfo{journal}{arXiv preprint arXiv:1910.08350} .
\bibitem[{Lee et~al.(2022)Lee, Lee and Park}]{lee2022augmentation}
\bibinfo{author}{Lee, N.}, \bibinfo{author}{Lee, J.}, \bibinfo{author}{Park, C.}, \bibinfo{year}{2022}.
\newblock \bibinfo{title}{Augmentation-free self-supervised learning on graphs}, in: \bibinfo{booktitle}{Proceedings of the AAAI Conference on Artificial Intelligence}, pp. \bibinfo{pages}{7372--7380}.
\bibitem[{Li et~al.(2023)Li, Guo, Ren, Yu, You and You}]{li2023explicit}
\bibinfo{author}{Li, H.}, \bibinfo{author}{Guo, Y.}, \bibinfo{author}{Ren, Z.}, \bibinfo{author}{Yu, F.R.}, \bibinfo{author}{You, J.}, \bibinfo{author}{You, X.}, \bibinfo{year}{2023}.
\newblock \bibinfo{title}{Explicit local coupling global structure clustering}.
\newblock \bibinfo{journal}{IEEE Transactions on Circuits and Systems for Video Technology} .
\bibitem[{Li et~al.(2021)Li, Kim and Wang}]{li2021beyond}
\bibinfo{author}{Li, S.}, \bibinfo{author}{Kim, D.}, \bibinfo{author}{Wang, Q.}, \bibinfo{year}{2021}.
\newblock \bibinfo{title}{Beyond low-pass filters: Adaptive feature propagation on graphs}, in: \bibinfo{booktitle}{Joint European Conference on Machine Learning and Knowledge Discovery in Databases}, \bibinfo{organization}{Springer}. pp. \bibinfo{pages}{450--465}.
\bibitem[{Li et~al.(2022)Li, Zhu, Cheng, Shan, Luo, Li and Qian}]{li2022finding}
\bibinfo{author}{Li, X.}, \bibinfo{author}{Zhu, R.}, \bibinfo{author}{Cheng, Y.}, \bibinfo{author}{Shan, C.}, \bibinfo{author}{Luo, S.}, \bibinfo{author}{Li, D.}, \bibinfo{author}{Qian, W.}, \bibinfo{year}{2022}.
\newblock \bibinfo{title}{Finding global homophily in graph neural networks when meeting heterophily}, in: \bibinfo{booktitle}{International Conference on Machine Learning}, \bibinfo{organization}{PMLR}. pp. \bibinfo{pages}{13242--13256}.
\bibitem[{Liang et~al.(2019)Liang, Huang and Wang}]{liang2019consistency}
\bibinfo{author}{Liang, Y.}, \bibinfo{author}{Huang, D.}, \bibinfo{author}{Wang, C.D.}, \bibinfo{year}{2019}.
\newblock \bibinfo{title}{Consistency meets inconsistency: A unified graph learning framework for multi-view clustering}, in: \bibinfo{booktitle}{2019 IEEE International Conference on Data Mining (ICDM)}, \bibinfo{organization}{IEEE}. pp. \bibinfo{pages}{1204--1209}.
\bibitem[{Lin et~al.(2023)Lin, Kang, Zhang and Tian}]{lin2023multi}
\bibinfo{author}{Lin, Z.}, \bibinfo{author}{Kang, Z.}, \bibinfo{author}{Zhang, L.}, \bibinfo{author}{Tian, L.}, \bibinfo{year}{2023}.
\newblock \bibinfo{title}{Multi-view attributed graph clustering}.
\newblock \bibinfo{journal}{IEEE Transactions on Knowledge and Data Engineering} \bibinfo{volume}{35}, \bibinfo{pages}{1872--1880}.
\bibitem[{Lingam et~al.(2021)Lingam, Iyer and Ragesh}]{lingam2021glam}
\bibinfo{author}{Lingam, V.}, \bibinfo{author}{Iyer, A.}, \bibinfo{author}{Ragesh, R.}, \bibinfo{year}{2021}.
\newblock \bibinfo{title}{Glam: Graph learning by modeling affinity to labeled nodes for graph neural networks}, in: \bibinfo{booktitle}{ICLR 2021 Workshop on Geometrical and Topological Representation Learning}.
\bibitem[{Liu et~al.(2022a)Liu, Jin, Pan, Zhou, Zheng, Xia and Yu}]{liu2022graph}
\bibinfo{author}{Liu, Y.}, \bibinfo{author}{Jin, M.}, \bibinfo{author}{Pan, S.}, \bibinfo{author}{Zhou, C.}, \bibinfo{author}{Zheng, Y.}, \bibinfo{author}{Xia, F.}, \bibinfo{author}{Yu, P.}, \bibinfo{year}{2022}a.
\newblock \bibinfo{title}{Graph self-supervised learning: A survey}.
\newblock \bibinfo{journal}{IEEE Transactions on Knowledge and Data Engineering} .
\bibitem[{Liu et~al.(2022b)Liu, Tu, Zhou, Liu, Song, Yang and Zhu}]{liu2022deep}
\bibinfo{author}{Liu, Y.}, \bibinfo{author}{Tu, W.}, \bibinfo{author}{Zhou, S.}, \bibinfo{author}{Liu, X.}, \bibinfo{author}{Song, L.}, \bibinfo{author}{Yang, X.}, \bibinfo{author}{Zhu, E.}, \bibinfo{year}{2022}b.
\newblock \bibinfo{title}{Deep graph clustering via dual correlation reduction}, in: \bibinfo{booktitle}{Proc. of AAAI}.
\bibitem[{Ma et~al.(2022)Ma, Liu, Shah and Tang}]{ma2022homophily}
\bibinfo{author}{Ma, Y.}, \bibinfo{author}{Liu, X.}, \bibinfo{author}{Shah, N.}, \bibinfo{author}{Tang, J.}, \bibinfo{year}{2022}.
\newblock \bibinfo{title}{Is homophily a necessity for graph neural networks?}, in: \bibinfo{booktitle}{International Conference on Learning Representations}.
\bibitem[{Pan and Kang(2021)}]{pan2021multi}
\bibinfo{author}{Pan, E.}, \bibinfo{author}{Kang, Z.}, \bibinfo{year}{2021}.
\newblock \bibinfo{title}{Multi-view contrastive graph clustering}.
\newblock \bibinfo{journal}{Advances in neural information processing systems} \bibinfo{volume}{34}, \bibinfo{pages}{2148--2159}.
\bibitem[{Pei et~al.(2019)Pei, Wei, Chang, Lei and Yang}]{pei2019geom}
\bibinfo{author}{Pei, H.}, \bibinfo{author}{Wei, B.}, \bibinfo{author}{Chang, K.C.C.}, \bibinfo{author}{Lei, Y.}, \bibinfo{author}{Yang, B.}, \bibinfo{year}{2019}.
\newblock \bibinfo{title}{Geom-gcn: Geometric graph convolutional networks}, in: \bibinfo{booktitle}{International Conference on Learning Representations}.
\bibitem[{Pei et~al.(2020)Pei, Wei, Chang, Lei and Yang}]{Pei2020Geom-GCN}
\bibinfo{author}{Pei, H.}, \bibinfo{author}{Wei, B.}, \bibinfo{author}{Chang, K.C.C.}, \bibinfo{author}{Lei, Y.}, \bibinfo{author}{Yang, B.}, \bibinfo{year}{2020}.
\newblock \bibinfo{title}{Geom-gcn: Geometric graph convolutional networks}, in: \bibinfo{booktitle}{International Conference on Learning Representations}.
\bibitem[{Platonov et~al.(2023)Platonov, Kuznedelev, Diskin, Babenko and Prokhorenkova}]{platonov2023critical}
\bibinfo{author}{Platonov, O.}, \bibinfo{author}{Kuznedelev, D.}, \bibinfo{author}{Diskin, M.}, \bibinfo{author}{Babenko, A.}, \bibinfo{author}{Prokhorenkova, L.}, \bibinfo{year}{2023}.
\newblock \bibinfo{title}{A critical look at the evaluation of gnns under heterophily: are we really making progress?}, in: \bibinfo{booktitle}{In The Eleventh International Conference on Learning Representations}.
\bibitem[{Pu et~al.(2021)Pu, Cao, Zhang, Dong and Chen}]{pu2021learning}
\bibinfo{author}{Pu, X.}, \bibinfo{author}{Cao, T.}, \bibinfo{author}{Zhang, X.}, \bibinfo{author}{Dong, X.}, \bibinfo{author}{Chen, S.}, \bibinfo{year}{2021}.
\newblock \bibinfo{title}{Learning to learn graph topologies}.
\newblock \bibinfo{journal}{Advances in Neural Information Processing Systems} \bibinfo{volume}{34}, \bibinfo{pages}{4249--4262}.
\bibitem[{Rozemberczki et~al.(2021)Rozemberczki, Allen and Sarkar}]{rozemberczki2021multi}
\bibinfo{author}{Rozemberczki, B.}, \bibinfo{author}{Allen, C.}, \bibinfo{author}{Sarkar, R.}, \bibinfo{year}{2021}.
\newblock \bibinfo{title}{Multi-scale attributed node embedding}.
\newblock \bibinfo{journal}{Journal of Complex Networks} \bibinfo{volume}{9}, \bibinfo{pages}{cnab014}.
\bibitem[{Tang et~al.(2009)Tang, Sun, Wang and Yang}]{tang2009social}
\bibinfo{author}{Tang, J.}, \bibinfo{author}{Sun, J.}, \bibinfo{author}{Wang, C.}, \bibinfo{author}{Yang, Z.}, \bibinfo{year}{2009}.
\newblock \bibinfo{title}{Social influence analysis in large-scale networks}, in: \bibinfo{booktitle}{Proceedings of the 15th ACM SIGKDD international conference on Knowledge discovery and data mining}, pp. \bibinfo{pages}{807--816}.
\bibitem[{Trivedi et~al.(2022)Trivedi, Lubana, Yan, Yang and Koutra}]{trivedi2022augmentations}
\bibinfo{author}{Trivedi, P.}, \bibinfo{author}{Lubana, E.S.}, \bibinfo{author}{Yan, Y.}, \bibinfo{author}{Yang, Y.}, \bibinfo{author}{Koutra, D.}, \bibinfo{year}{2022}.
\newblock \bibinfo{title}{Augmentations in graph contrastive learning: Current methodological flaws \& towards better practices}, in: \bibinfo{booktitle}{Proceedings of the ACM Web Conference 2022}, pp. \bibinfo{pages}{1538--1549}.
\bibitem[{Veli{\v{c}}kovi{\'c} et~al.(2019)Veli{\v{c}}kovi{\'c}, Cucurull, Casanova, Romero, Li{\`o} and Bengio}]{velivckovic2018graph}
\bibinfo{author}{Veli{\v{c}}kovi{\'c}, P.}, \bibinfo{author}{Cucurull, G.}, \bibinfo{author}{Casanova, A.}, \bibinfo{author}{Romero, A.}, \bibinfo{author}{Li{\`o}, P.}, \bibinfo{author}{Bengio, Y.}, \bibinfo{year}{2019}.
\newblock \bibinfo{title}{Graph attention networks}, in: \bibinfo{booktitle}{International Conference on Learning Representations}.
\bibitem[{Wang et~al.(2019)}]{wang2019attributed}
\bibinfo{author}{Wang, C.}, et~al., \bibinfo{year}{2019}.
\newblock \bibinfo{title}{Attributed graph clustering: A deep attentional embedding approach}, in: \bibinfo{booktitle}{IJCAI}.
\bibitem[{Wang et~al.(2020)Wang, Yang and Liu}]{wang2020gmc}
\bibinfo{author}{Wang, H.}, \bibinfo{author}{Yang, Y.}, \bibinfo{author}{Liu, B.}, \bibinfo{year}{2020}.
\newblock \bibinfo{title}{{GMC}: {G}raph-based multi-view clustering}.
\newblock \bibinfo{journal}{{IEEE} Transactions on Knowledge and Data Engineering} \bibinfo{volume}{32}, \bibinfo{pages}{1116--1129}.
\bibitem[{Wang et~al.(2021a)Wang, Mou, Wang, Xiao, Ju, Shi and Xie}]{wang2021graph}
\bibinfo{author}{Wang, R.}, \bibinfo{author}{Mou, S.}, \bibinfo{author}{Wang, X.}, \bibinfo{author}{Xiao, W.}, \bibinfo{author}{Ju, Q.}, \bibinfo{author}{Shi, C.}, \bibinfo{author}{Xie, X.}, \bibinfo{year}{2021}a.
\newblock \bibinfo{title}{Graph structure estimation neural networks}, in: \bibinfo{booktitle}{Proceedings of the Web Conference 2021}, pp. \bibinfo{pages}{342--353}.
\bibitem[{Wang et~al.(2022)Wang, Jin, Wang, He and Huang}]{wang2022powerful}
\bibinfo{author}{Wang, T.}, \bibinfo{author}{Jin, D.}, \bibinfo{author}{Wang, R.}, \bibinfo{author}{He, D.}, \bibinfo{author}{Huang, Y.}, \bibinfo{year}{2022}.
\newblock \bibinfo{title}{Powerful graph convolutional networks with adaptive propagation mechanism for homophily and heterophily}, in: \bibinfo{booktitle}{Proceedings of the AAAI Conference on Artificial Intelligence}, pp. \bibinfo{pages}{4210--4218}.
\bibitem[{Wang et~al.(2021b)Wang, Wu, Zhang, Zhou, Chen and Liu}]{wang2021multi}
\bibinfo{author}{Wang, T.}, \bibinfo{author}{Wu, J.}, \bibinfo{author}{Zhang, Z.}, \bibinfo{author}{Zhou, W.}, \bibinfo{author}{Chen, G.}, \bibinfo{author}{Liu, S.}, \bibinfo{year}{2021}b.
\newblock \bibinfo{title}{Multi-scale graph attention subspace clustering network}.
\newblock \bibinfo{journal}{Neurocomputing} \bibinfo{volume}{459}, \bibinfo{pages}{302--314}.
\bibitem[{Wang et~al.(2017)Wang, Cui, Wang, Pei, Zhu and Yang}]{wang2017community}
\bibinfo{author}{Wang, X.}, \bibinfo{author}{Cui, P.}, \bibinfo{author}{Wang, J.}, \bibinfo{author}{Pei, J.}, \bibinfo{author}{Zhu, W.}, \bibinfo{author}{Yang, S.}, \bibinfo{year}{2017}.
\newblock \bibinfo{title}{Community preserving network embedding}, in: \bibinfo{booktitle}{Proceedings of the AAAI conference on artificial intelligence}.
\bibitem[{Wong et~al.(2019)Wong, Han, Fang, Zhan and Wen}]{wong2019clustering}
\bibinfo{author}{Wong, W.K.}, \bibinfo{author}{Han, N.}, \bibinfo{author}{Fang, X.}, \bibinfo{author}{Zhan, S.}, \bibinfo{author}{Wen, J.}, \bibinfo{year}{2019}.
\newblock \bibinfo{title}{Clustering structure-induced robust multi-view graph recovery}.
\newblock \bibinfo{journal}{IEEE Transactions on Circuits and Systems for Video Technology} \bibinfo{volume}{30}, \bibinfo{pages}{3584--3597}.
\bibitem[{Wu et~al.(2019)Wu, Souza, Zhang, Fifty, Yu and Weinberger}]{wu2019simplifying}
\bibinfo{author}{Wu, F.}, \bibinfo{author}{Souza, A.}, \bibinfo{author}{Zhang, T.}, \bibinfo{author}{Fifty, C.}, \bibinfo{author}{Yu, T.}, \bibinfo{author}{Weinberger, K.}, \bibinfo{year}{2019}.
\newblock \bibinfo{title}{Simplifying graph convolutional networks}, in: \bibinfo{booktitle}{International conference on machine learning}, \bibinfo{organization}{PMLR}. pp. \bibinfo{pages}{6861--6871}.
\bibitem[{Xie et~al.(2023)Xie, Chen, Kang and Peng}]{CGC}
\bibinfo{author}{Xie, X.}, \bibinfo{author}{Chen, W.}, \bibinfo{author}{Kang, Z.}, \bibinfo{author}{Peng, C.}, \bibinfo{year}{2023}.
\newblock \bibinfo{title}{Contrastive graph clustering with adaptive filter}.
\newblock \bibinfo{journal}{Expert Systems with Applications} \bibinfo{volume}{219}, \bibinfo{pages}{119645}.
\bibitem[{Xu et~al.(2018)Xu, Li, Tian, Sonobe, Kawarabayashi and Jegelka}]{xu2018representation}
\bibinfo{author}{Xu, K.}, \bibinfo{author}{Li, C.}, \bibinfo{author}{Tian, Y.}, \bibinfo{author}{Sonobe, T.}, \bibinfo{author}{Kawarabayashi, K.i.}, \bibinfo{author}{Jegelka, S.}, \bibinfo{year}{2018}.
\newblock \bibinfo{title}{Representation learning on graphs with jumping knowledge networks}, in: \bibinfo{booktitle}{International conference on machine learning}, \bibinfo{organization}{PMLR}. pp. \bibinfo{pages}{5453--5462}.
\bibitem[{You et~al.(2020)You, Chen, Sui, Chen, Wang and Shen}]{you2020graph}
\bibinfo{author}{You, Y.}, \bibinfo{author}{Chen, T.}, \bibinfo{author}{Sui, Y.}, \bibinfo{author}{Chen, T.}, \bibinfo{author}{Wang, Z.}, \bibinfo{author}{Shen, Y.}, \bibinfo{year}{2020}.
\newblock \bibinfo{title}{Graph contrastive learning with augmentations}.
\newblock \bibinfo{journal}{Advances in Neural Information Processing Systems} \bibinfo{volume}{33}, \bibinfo{pages}{5812--5823}.
\bibitem[{Zhang et~al.(2023)Zhang, Li, Qiu, Tang, Wen, Zha and Wen}]{zhang2023efficient}
\bibinfo{author}{Zhang, H.}, \bibinfo{author}{Li, S.}, \bibinfo{author}{Qiu, J.}, \bibinfo{author}{Tang, Y.}, \bibinfo{author}{Wen, J.}, \bibinfo{author}{Zha, Z.}, \bibinfo{author}{Wen, B.}, \bibinfo{year}{2023}.
\newblock \bibinfo{title}{Efficient and effective nonconvex low-rank subspace clustering via svt-free operators}.
\newblock \bibinfo{journal}{IEEE Transactions on Circuits and Systems for Video Technology} .
\bibitem[{Zhang et~al.(2019a)Zhang, Liu, Li and Wu}]{zhang2019attributed}
\bibinfo{author}{Zhang, X.}, \bibinfo{author}{Liu, H.}, \bibinfo{author}{Li, Q.}, \bibinfo{author}{Wu, X.M.}, \bibinfo{year}{2019}a.
\newblock \bibinfo{title}{Attributed graph clustering via adaptive graph convolution}, in: \bibinfo{booktitle}{IJCAI}.
\bibitem[{Zhang and Zitnik(2020)}]{zhang2020gnnguard}
\bibinfo{author}{Zhang, X.}, \bibinfo{author}{Zitnik, M.}, \bibinfo{year}{2020}.
\newblock \bibinfo{title}{Gnnguard: Defending graph neural networks against adversarial attacks}.
\newblock \bibinfo{journal}{Advances in neural information processing systems} \bibinfo{volume}{33}, \bibinfo{pages}{9263--9275}.
\bibitem[{Zhang et~al.(2019b)Zhang, Pal, Coates and Ustebay}]{zhang2019bayesian}
\bibinfo{author}{Zhang, Y.}, \bibinfo{author}{Pal, S.}, \bibinfo{author}{Coates, M.}, \bibinfo{author}{Ustebay, D.}, \bibinfo{year}{2019}b.
\newblock \bibinfo{title}{Bayesian graph convolutional neural networks for semi-supervised classification}, in: \bibinfo{booktitle}{Proceedings of the AAAI conference on artificial intelligence}, pp. \bibinfo{pages}{5829--5836}.
\bibitem[{Zhao et~al.(2023)Zhao, Shen, Chen, Liang, Chen and Zhou}]{zhao2023self}
\bibinfo{author}{Zhao, X.}, \bibinfo{author}{Shen, Q.}, \bibinfo{author}{Chen, Y.}, \bibinfo{author}{Liang, Y.}, \bibinfo{author}{Chen, J.}, \bibinfo{author}{Zhou, Y.}, \bibinfo{year}{2023}.
\newblock \bibinfo{title}{Self-completed bipartite graph learning for fast incomplete multi-view clustering}.
\newblock \bibinfo{journal}{IEEE Transactions on Circuits and Systems for Video Technology} .
\bibitem[{Zhou et~al.(2004)Zhou, Bousquet and Lal}]{zhou2004learning}
\bibinfo{author}{Zhou, D.}, \bibinfo{author}{Bousquet, O.}, \bibinfo{author}{Lal, T.N.}, \bibinfo{year}{2004}.
\newblock \bibinfo{title}{Learning with local and global consistency}, in: \bibinfo{booktitle}{Advances in Neural Information Processing Systems 16: Proceedings of the 2003 Conference}, \bibinfo{organization}{MIT Press}. p. \bibinfo{pages}{321}.
\bibitem[{Zhu et~al.(2022)}]{zhu2022collaborative}
\bibinfo{author}{Zhu, P.}, et~al., \bibinfo{year}{2022}.
\newblock \bibinfo{title}{Collaborative decision-reinforced self-supervision for attributed graph clustering}.
\newblock \bibinfo{journal}{IEEE Transactions on Neural Networks and Learning Systems} .
\bibitem[{Zhu et~al.(2021)Zhu, Xu, Yu, Liu, Wu and Wang}]{zhu2021graph}
\bibinfo{author}{Zhu, Y.}, \bibinfo{author}{Xu, Y.}, \bibinfo{author}{Yu, F.}, \bibinfo{author}{Liu, Q.}, \bibinfo{author}{Wu, S.}, \bibinfo{author}{Wang, L.}, \bibinfo{year}{2021}.
\newblock \bibinfo{title}{Graph contrastive learning with adaptive augmentation}, in: \bibinfo{booktitle}{WWW}.

\end{thebibliography}
\end{document}